# Lane Detection Model Based on Spatio-Temporal Network With Double Convolutional Gated Recurrent Units

Jiyong Zhang, Tao Deng, Fei Yan, and Wenbo Liu

*Abstract*—Lane detection is one of the indispensable and key elements of self-driving environmental perception. Many lane detection models have been proposed, solving lane detection under challenging conditions, including intersection merging and splitting, curves, boundaries, occlusions and combinations of scene types. Nevertheless, lane detection will remain an open problem for some time to come. The ability to cope well with those challenging scenes impacts greatly the applications of lane detection on advanced driver assistance systems (ADASs). In this paper, a spatio-temporal network with double Convolutional Gated Recurrent Units (ConvGRUs) is proposed to address lane detection in challenging scenes. Both of ConvGRUs have the same structures, but different locations and functions in our network. One is used to extract the information of the most likely low-level features of lane markings. The extracted features are input into the next layer of the end-to-end network after concatenating them with the outputs of some blocks. The other one takes some continuous frames as its input to process the spatio-temporal driving information. Extensive experiments on the large-scale TuSimple lane marking challenge dataset and Unsupervised LLAMAS dataset demonstrate that the proposed model can effectively detect lanes in the challenging driving scenes. Our model can outperform the state-of-the-art lane detection models.

## I. INTRODUCTION

ADVANCED driver assistance systems (ADASs) have become a hot topic in the current computer vision and autonomous driving research. The main bottleneck of autonomous driving is the environmental perception problem [1]. Self-driving itself is a very complex problem due to the changes in autonomous driving environments. Environmental changes include many factors, and each one is a challenging subtask for an ADAS, such as road detection, lane detection, vehicle detection, pedestrian detection, drowsiness detection, collision avoidance and traffic sign detection [2].

Manuscript received August 10, 2020; revised January 11, 2021; accepted February 12, 2021. This work was supported in part by the National Natural Science Foundation of China under Grant 61873215, in part by the Fundamental Research Funds for the Central Universities under Grant 2682019CX59, in part by the Key Program for International S&T Cooperation of Sichuan Province under Grant 2019YFH0097, and in part by the Sichuan Science and Technology Program under Grant 2018GZ0008 and Grant 2020JDRC0031. The Associate Editor for this article was L. M. Bergasa. *(Corresponding author: Tao Deng.)*

The authors are with the School of Information Science and Technology, Southwest Jiaotong University, Chengdu 611756, China (e-mail: zhangjiyong-com@163.com; tdeng@swjtu.edu.cn; fyan@swjtu.edu.cn; l2571630192@163.com).

Digital Object Identifier 10.1109/TITS.2021.3060258

**Lane perception**, as one of the two primary related technologies given by [3], **is a crucial step towards a fully autonomous car, and it can help a car to place in itself among the lanes and perceive its surroundings when driving on roads.**

To tackle such changes, many lane detection techniques have been developed. They are divided into traditional method-based techniques and deep learning-based technologies. By studying many relevant literatures, we know that traditional methods are often applied in situations in which the changes of road scenes are not obvious, and the advantage of a deep learning method is that it can overcome the problem in which scenario changes generally cause traditional methods to fail.

The contents of lane detection have the following characteristics. First, the lane lines themselves exist in different scenes in different periods on different sections of a road. For example, they may include both shadows and dashed lanes at time $T_1$. Then at time $T_2$, the situation may contain shadows and unclear lanes. However, at time $T_3$, the road scenes may comprise curved lines, broken lines, split lines, and so on. The changing environment mentioned above is hard to circumvent. Therefore, although many lane detection approaches have been proposed, lane detection in challenging scenes, still remains unresolved in fact, as shown in Fig. 1. This paper is inspired by the following two points. One point is the relationship between visual perception and memory, which is taken from Brodmann's brain map and the relative theories [5]–[7]. The other point is the successful application of semantic segmentation to image processing [8]–[11].

In this paper, we regard lane detection as a segmentation problem according to the second point above. The first point enlightens us to consider adding one Convolutional Gated Recurrent Unit (ConvGRU) [12] in the low-level feature extraction phase in our end-to-end neural network, which then concatenate the output of the added ConvGRU unit with some intermediate results and use them as an input vector in the next layer. In order to detect lanes in some challenging scenes, we design an end-to-end neural network with double ConvGRU units to cope with it. To summarize, the contributions of this work are the following:

- An encoder-decoder network with double ConvGRUs is proposed in this work.
- Both the spatial and temporal driving information are considered in the lane detection problem.

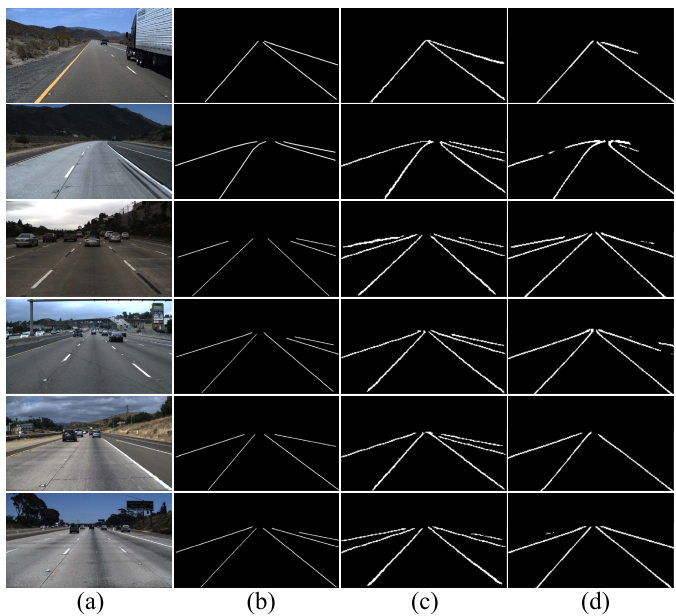

Fig. 1. Lane detection in challenging situations. (a) Different driving scenes. (b) Ground truth. (c) Lane detection results of the proposed method. (d) Lane detection results of the U-Net_ConvLSTM [4].

- A Front ConvGRU is used to extract more accurate information on the low-level features.
- Our lane detection model can detect lane lines in challenging scenes such as curves, junctions, occlusions due to vehicles and so on.

The remainder of this paper is organized as follows. Section II reviews the related lane detection research work. Section III describes the proposed network. Section IV reports the designed experiments and analyzes the results. The effects of our proposed model are also investigated in this section. Section V discusses some problems encountered during the training and testing processes. Section VI concludes our work.

## II. RELATED WORK

In the past two decades, a number of studies on lane detection have been published. These methods are mainly based on classical image processing and deep learning methods. In this section, some existing works on lane detection and prediction are reviewed briefly.

### A. Lane Detection Based on Classical Image Processing

Although roads are structured, the lane lines on roads vary for a moving car. With respect to traditional lane detection, Borkar et al. [13] utilized the Inverse Perspective Transformation (IPM) to transform images, then used a Random Sample Consensus (RANSAC) algorithm to remove interference, and last, adopted the Kalman filter to complete the lane line prediction. Aiming at the multiple lane problem, a Bayes filtering method based on multi-objects was studied to predict for lane markings in bad conditions [14]. Hur et al. [15] posed a CRFs-based method to deal with the problem that complex urban lane situations included parallel lines but no parallel scenes and proved the validity of their designed algorithm through related experiments. Tan et al. [16] devised the Improved River Flow (IRF) and RANSAC-based method to address the curves in the changing conditions stated above. The IRF was used to obtain feature points, and the RANSAC calculated the curvatures that were utilized to fit the curved lane lines in the postprocessing phase. From the color-information point of view, Chiu et al. [17] merely addressed lane detection as a classification problem. They extracted color-based segmentation information with the help of probability knowledge, and applied a quadratic function to find out lane boundaries. Finally, both the extracted information and boundaries were used together to detect lane lines. To handle the challenge of inner-city scenes without distinguishing ego-lane scenes, Kuhnl et al. [18] designed an approach based on confidence maps. The confidence maps were used as the basic classifiers to produce spatial ray features which were calculated and used to determine the existence of ego-lanes. Considering both the light intensity and width of lane markings, Liu et al. [19] combined a local threshold segmentation algorithm and Hough transform with a few subsidiary prerequisites to detect lane markings.

### B. Lane Detection Based on Deep Learning

Deep learning-based approaches have been the mainstream in recent years. These methods can be broadly divided into two categories: classification-based lane detection and semantic segmentation-based lane detection.

Considering the spatial structures of lane markings, Li et al. constructed a deep neural network that consisted of a multitask deep convolutional network and a recurrent neural network [20]. The former detected the target and its geometric attributes, including its location and orientation. The latter dealt with the spatial visual cues distributed in an object. Last, their outputs were combined together to produce lane prediction and recognize the lane markings. Kim et al. [21] mixed a simple CNN framework and the RANSAC algorithm to detect lanes and lane markings. For simple road scenes, they only used the RANSAC algorithm to check the lane markings. For complex scenes that included roadside trees, fences, etc., they adopted the CNN to measure the lane markings after the RANSAC processing. Although semantic segmentation is a common lane detection method, Ghafoorian et al. [22] thought that some lanes can not be treated as pixel-wise classification in semantic segmentation, hence, they proposed the EL-GAN network to conduct lane detection. The EL-GAN network can produce a structure-preserving output that stabilized the training process and further improved the model performance according to their experimental results. When running on the road, the first aim of the autonomous driving was to position its own ego-lanes and side lanes. Chougule et al. [23] formed it as a detection and classification problem, and they combined a CNN network with regression approaches to conduct training according to that idea. Their experiments showed that the trained model cooperatively worked well at recognizing thin and elongated boundaries, especially occlusions caused by vehicles. Chen et al. [24] proposed a different method using common approaches. They constructed an end-to-end network that output a steering angle using raw image as the input, and the steering angle can be directly used to operate the car and keep the self-driving car in its lane. A study by Lee et al. [25], established a VPGNet framework to predict vanishing points by training their proposed neural network, and it further detected and classified lanes and road markings in the following challenging scenes: no rain, rain, heavy rain, and night. Due to complex scenes stated previously, there was no existing benchmark dataset that can be used when they evaluated the validity and accuracy of their proposed model, therefore, the authors designed their own dataset that consisted of 20,000 images with 17 classes. Finally, the experiments showed the validity of their proposed algorithm.

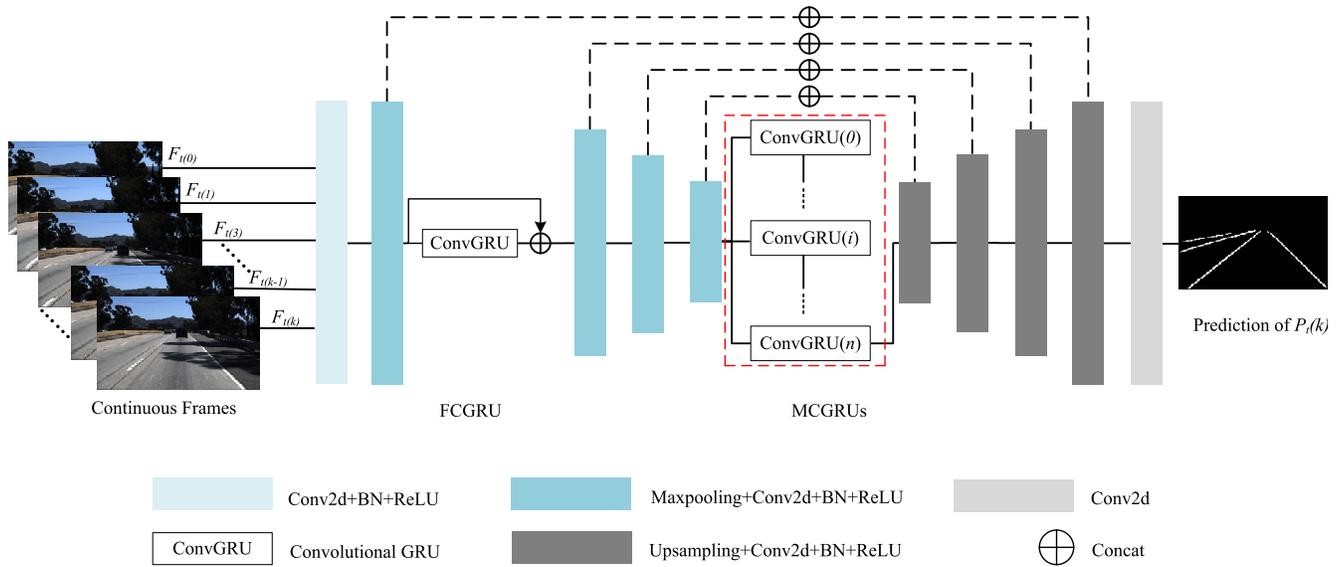

Fig. 2. Architecture overview. FCGRU: Front ConvGRU. MCGRUs: Middle ConvGRUs.

Increasingly more deep learning methods based on semantic segmentation have been applied for lane detection. Yeongmin et al. [26] proposed a lane detection framework (named PINet) that includes several branches, which roughly divided the whole lane detection task into three parts. The first was to extract the features of input images with the help of hourglass network [27], and produced three results, including confidence, offset, and feature. The second was to use confidence and offset to predict the exact points, and distinguished the predicted exact points into each instance. The last is to generate the smooth lane by applying the post processing technology. Neven et al. regarded lane detection as an instance problem, and built a LaneNet framework that consisted of two subneural networks, including an embedding branch and a segmentation binary branch [28]. The former generated a pixel embedding, which was further processed using a discrimination operation in order to obtain clear and accurate pixels for each lane. The latter produced a binary mask feature map. Both of them were cooperatively and orderly clustered to produce the final result of LaneNet. Furthermore, the HNet was constructed to train the parameters for the purpose of fitting the curves when producing the final result.

Qin et al. [29] regarded lane detection as a row selection problem with global information. Row-based selecting helped them reduce the corresponding computational cost, and global information helped them handle the challenging scenes. For lane detection in the challenging scene of a weak visual appearance, Du et al. [30], further highlighted the feature representation capability of the CNN via discussing different settings in their proposed multiple encoder-decoder networks. In semantic segmentation, the spatial relationships between pixels were considered the main point by Pan et al. [31]. They modified a CNN and constructed the Spatial CNN (SCNN) framework to train a model and let the model detect the shapes and structures of lane lines. The continuity of information between multiple frames was considered by Zou et al. [4], a novel hybrid deep architecture that combined CNN and the recurrent neural network (RNN) was proposed to detect lane lines.

III. PROPOSED NETWORK

In this section, we introduce the proposed network that fuses an end-to-end CNN and two ConvGRUs in order to detect lane lines in challenging driving scenes.

A. Proposed Network Description

Our network includes encoder and decoder networks. The architecture overview of our model is shown in Fig. 2. Each of the convolution blocks includes 2-D convolutions, batch normalization [32], maxpooling, upsampling and an ReLU [33] activation function. In the encoder phase, we add a ConvGRU after the second convolution block. In the human brain, we guess that the low-level features (such as colors, shapes, boundaries and so on) should be correspondingly remembered. Some memory related brain regions may participate in the low-level feature extraction phase [34], [35]. Therefore, one ConvGRU is used in the encoder phase for memorizing and learning the low-level features. More descriptions are stated in Section III-B. The output of the encoder network is input into the multiple ConvGRUs, which are added to make full use of spatio-temporal information of continuous driving images. The detailed contents are also expatiated in Section III-B. Finally, the decoder network is responsible for decoding the content as an image that has the same size as the original input images. Note that there is a skip connection operation similar to that of the U-Net [36], which works between the encoder and decoder phase in order to provide more information to obtain more accurate predictions of lane lines.

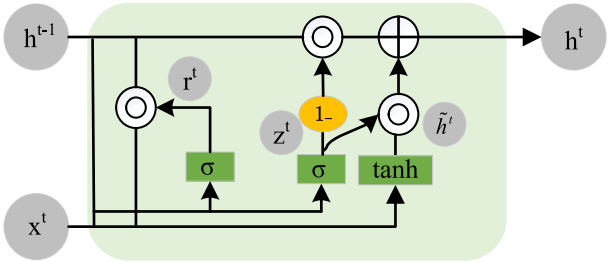

Fig. 3. Internal structure of ConvGRU.

### B. End-to-End CNN and ConvGRU

The end-to-end CNN and ConvGRU have their own strengths. Learning more about those advantages is foundational and necessary for cooperatively using the end-to-end CNN and ConvGRU to build a model for solving the lane detection problem.

*1) CNN:* The ability of the end-to-end CNN has been proved by many existing studies [12], [33], [37], [38], [39]. The end-to-end CNN architecture is better at extracting features by convolution operations and pooling operations, especially those of regular objects. Furthermore, the end-to-end network can obtain images with the same size as the inputs by convolution operations and upsampling operations. Road lanes have regular markings, and their attributes include their location, color, width, length, and so on. Therefore, an end-to-end CNN is more suitable for the feature extraction of lane detection.

*2) ConvGRU:* The ConvGRU [12], [40], which has the same or even slightly higher performance than the RNN or ConvLSTM, has a simpler structure and less memory consuming [41]. The acquisition of the current state is not only related to the current input, but also to some extent related to the previous moment. The internal structure of the ConvGRU is shown in Fig. 3. It is a nonlinear time-series model that has the computing complexity and time memory abilities when processing spatio-temporal information. The convolutional operations in the ConvGRU give it a stronger feature learning ability by making the internal coefficients of the proposed model no longer fixed. The changed coefficients allow the model to better fit and learn the current context when extracting features. Therefore, the last output of the ConvGRU is the most likely information to be remembered. The process reflects that the ConvGRU has the learning and memory abilities.

The computational Formulas of the ConvGRU are shown in the following:

$$z^t = \sigma(W_z^t * x^t + U_z^t * h^{t-1} + b_z) \quad (1)$$

$$r^t = \sigma(W_r^t * x^t + U_r^t * h^{t-1} + b_r) \quad (2)$$

$$\tilde{h}^t = \tanh(W^t * x^t + U^t * (r^t \odot h^{t-1}) + b) \quad (3)$$

$$h^t = (1 - z^t)h^{t-1} + z^t \tilde{h}^t \quad (4)$$

In the above Formulas, * means the convolution operation. $z^t$ is the update gate of layer $l$ at time $t$, which actually decides to what extent to update when producing the final result $h^t$ at layer $l$ in Formula 4. $r^t$ is the reset gate at time $t$. Although the driving environment conditions change over time, it is not appropriate to leave out all previous information [4], [12]. Because the current scene is related to its previous moment. At this point, $r^t$ is used to control how much feature information should be forgotten by an element-wise multiplication operation with the previous hidden state information when current candidate hidden information is calculated. $\odot$ represents the element-wise multiplication operation. $\tilde{h}^t$ is the current candidate hidden representation that is calculated using activation function $tanh$. It is used to multiply an update gate at layer $l$. To be exact, most of the information of $h^t$ comes from $\tilde{h}^t$. $h^{t-1}$ is a previous hidden-state representation of layer $l$, which participates in the whole computing process of building the final feature $h^t$. In addition, $W_r^t$, $W_z^t$, and $W$, $U_r^t$, $U_z^t$, $U$ denote different convolution kernel variables that have different subscripts to represent the different stages. They are used when calculating the corresponding reset gate, update gate and current candidate hidden representation, respectively. $b_z$, $b_r$, and $b$ are the biases used for adjusting the corresponding outputs at different stages. $x^t$ is an input feature vector. $\sigma(\cdot)$ represents the sigmoid operation and $tanh(\cdot)$ represents the hyperbolic tangent nonlinearities.

Based on the advantages and theories of both the ConvGRU and the end-to-end CNN, and the corresponding conjecture described above, we choose to put two ConvGRUs in an end-to-end CNN at different locations when designing our model, which can be seen in Fig. 2.

*3) FCGRU and MCGRUs:* The FCGRU shown in Fig. 2 denotes Front ConvGRU unit located behind the second convolution block in the encoder phase. At this time, the information of the current frame is processed in the whole calculation process of the FCGRU. In this case, the variable $t$ equals 1, only the current frame $x^t$ is fed into the FCGRU. The variables $x^t$ and $h^{t-1}$ have the same size $B \times C \times H \times W$ ($B$ =batch size = 6, $C$ = 48, $H$ = 64, $W$ = 128). Before using the FCGRU for the first time to extract the low-level features of the current frame, the variable $h^{t-1}$ should be initialized based on the size of the variable $x^t$. After that, they are used to calculate the corresponding output $h^t$ according to Formulas 1 to 4 in combination with the initialized weights. Both the learning and short-term memory abilities from the internal nonlinear attribute of ConvGRU are fully utilized to extract the low-level features at the beginning phase of the encoding network.

The reason why we implement an FCGRU unit after conv2_2 is based on the relationship between the low-level features and high-level features implied by Brodmann's relative theories [5]–[7] and their functions in predicting fixation [42]. It has been proven that there is a certain relationship between visual perception and memory [34], [35]. We hypothesize that the brain memory areas (such as the hippocampus) can aid the primary visual cortex (such as the LGN and, V1). Before the objects are recognized, the low-level features of objects may also be recognized. The FCGRU can memorize the features of lane lines more accurately. Some corresponding samples of the abstracted feature maps are shown in Fig. 4.

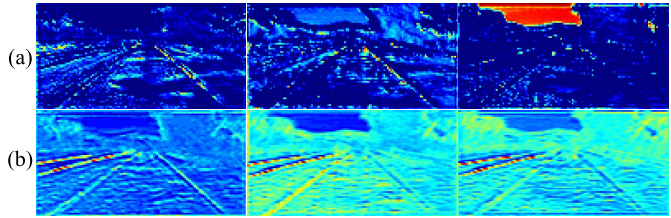

Fig. 4. Low-level features of lanes. (a) Results of conv2_2. (b) Results of the FCGRU.

The feature maps of conv2_2 (Fig. 4 (a)) and the outputs of the FCGRU (Fig. 4 (b)) are shown in Fig. 4. It is not difficult to see the difference between them. In Fig. 4 (a), the CNN has the feature extraction ability from images and obtains a large number of features about the target lane lines using the repeated convolution operations and maxpooling operations. However, the generated low-level features from the CNN may be not clear for the lane detection task in the initial stage of the encoding network. There is plenty of interference signals in them. There is no doubt that a large amount of interference information is likely to affect the training model of the neural network, especially its weight parameters. It may further affect the generation of the final lane detection results. In Fig. 4 (b), the low-level features of the lane lines are obvious. There are few traces surrounding them.

The Middle ConvGRUs (MCGRUs) shown in Fig. 2 is implemented between the encoder and decoder phases. It has a different function than the FCGRU. It takes $K$ continuous frames as its input and gets one output corresponding to the $K^{th}$ frame. The input frames can provide a wealth of spatio-temporal information about lane lines. As a matter of fact, it is with this point in mind that the MCGRUs located in front of the decoder network are used to extract the high-level features by making the best of the long-short-term memory ability. During this feature extraction process, when we take 5 ($K = 5$) continuous frames into the MCGRUs, it implies the variable $x^t$ represents different features of different frames at different time ($t = 1, 2, 3, 4, 5$). At this point, the variables $x^t$ and $h^{t-1}$ have the same size $B \times C \times H \times W$ ($B$ =batch size = 6, $C = 128$, $H = 8$, $W = 16$). When the variable $t$ equals to 1, the calculation process is similar to that of the FCGRU unit. At this moment, $x^t$ and $h^{t-1}$ use the same content in the process of calculating low-level features. When the variable $t$ equals to 2, $x^t$ represents the features of the second frame, and $h^{t-1}$ represents the output of the previous frame. Until $t$ equals to 5, the corresponding features of lane lines in the $5^{th}$ frame is predicted. Obviously, $x^t$ and $h^{t-1}$ are truly different from their previous moments in the process of calculating and producing the features by using the MCGRUs. Through the analysis of Formulas 1 - 4 and the comparison of the above features in Fig. 4, it follows that the ConvGRUs can accurately and effectively obtain the lane features by calculating the variables in the above Formulas.

TABLE I
NETWORK ARCHITECTURE PARAMETERS

| Layer | Input | Output | Kernel | Stride | Pad | Activation |
|---|---|---|---|---|---|---|
| conv1_1 | 256×128×3 | 256×128×32 | 3×3 | 1 | 1 | ReLU |
| conv1_2 | 256×128×32 | 256×128×32 | 3×3 | 1 | 1 | ReLU |
| Maxpool2d | 256×128×32 | 128×64×32 | 2×2 | 2 | 0 | |
| conv2_1 | 128×64×32 | 128×64×48 | 3×3 | 1 | 1 | ReLU |
| conv2_2 | 128×64×48 | 128×64×48 | 3×3 | 1 | 1 | ReLU |
| FCGRU | | | | | | |
| Maxpool2d | 128×64×96 | 64×32×96 | 2×2 | 2 | 0 | |
| conv3_1 | 64×32×96 | 64×32×64 | 3×3 | 1 | 1 | ReLU |
| conv3_2 | 64×32×64 | 64×32×64 | 3×3 | 1 | 1 | ReLU |
| Maxpool2d | 64×32×64 | 32×16×64 | 2×2 | 2 | 0 | |
| conv4_1 | 32×16×64 | 32×16×128 | 3×3 | 1 | 1 | ReLU |
| conv4_2 | 32×16×128 | 32×16×128 | 3×3 | 1 | 1 | ReLU |
| Maxpool2d | 32×16×128 | 16×8×128 | 2×2 | 2 | 0 | |
| conv5_1 | 16×8×128 | 16×8×128 | 3×3 | 1 | 1 | ReLU |
| conv5_2 | 16×8×128 | 16×8×128 | 3×3 | 1 | 1 | ReLU |
| MCGRUs | | | | | | |
| Upsampling | 16×8×128 | 32×16×128 | 2×2 | 2 | 0 | |
| conv6_1 | 32×16×256 | 32×16×128 | 3×3 | 1 | 1 | ReLU |
| conv6_2 | 32×16×128 | 32×16×128 | 3×3 | 1 | 1 | ReLU |
| Upsampling | 32×16×128 | 64×32×128 | 2×2 | 2 | 0 | |
| conv7_1 | 64×32×192 | 64×32×64 | 3×3 | 1 | 1 | ReLU |
| conv7_2 | 64×32×64 | 64×32×64 | 3×3 | 1 | 1 | ReLU |
| Upsampling | 64×32×64 | 128×64×64 | 2×2 | 2 | 0 | |
| conv8_1 | 128×64×112 | 128×64×48 | 3×3 | 1 | 1 | ReLU |
| conv8_2 | 128×64×48 | 128×64×48 | 3×3 | 1 | 1 | ReLU |
| Upsampling | 128×64×48 | 256×128×48 | 2×2 | 2 | 0 | |
| conv9_1 | 256×128×80 | 256×128×16 | 3×3 | 1 | 1 | ReLU |
| conv9_2 | 256×128×16 | 256×128×16 | 3×3 | 1 | 1 | ReLU |
| conv10_1 | 256×128×16 | 256×128×2 | 3×3 | 1 | 1 | ReLU |

\* FCGRU: Front ConvGRU. MCGRUs: Middle ConvGRUs.

### C. Implementation Details

*1) Network Details:* The proposed network is used for the lane detection task. At the beginning, $K$ continuous image frames ($K = 5$ in our work) are input into the convolution blocks of the encoder network. The outputs of conv2_2 are fed into the FCGRU. Then, the outputs of the encoder network are taken into the MCGRUs. More useful and accurate features are further to be extracted by dealing with spatio-temporal signals. In a word, the encoder network is primarily responsible for extracting features converting larger images into the specified sized images, increasing the number of channels and iteratively adjusting the weights in order to minimize the loss.

The outputs of the MCGRUs are input into the decoder network that rebuilds and highlights the features using deconvolution and upsampling operations at pixel-level granularity. Finally, the lane prediction result $P_t(k)$ with the same size as the original input frame is obtained after the decoder phase. The detailed parameters of the entire network are listed in Table I.

*2) Training Details:* After constructing the proposed network, it is trained by repeatedly updating the weight parameters and loss based on the deviation between the ground truth and output from the proposed network. We set the input size to $128 \times 256$, the batch size to 6 due to the limitation of our GPU and the learning rate to 0.001. The RAdam [43] optimizer is used in our work. Compared with the performance of the SGD [44] and Adam [45], the corresponding loss of RAdam is the lowest in our experiments. The cross-entropy loss function is used to calculate the loss in our work. The model is trained on a platform with an Intel Core i7-6800k CPU, 64GB of RAM, and one NVIDIA TITAN Xp 12GB GPU.

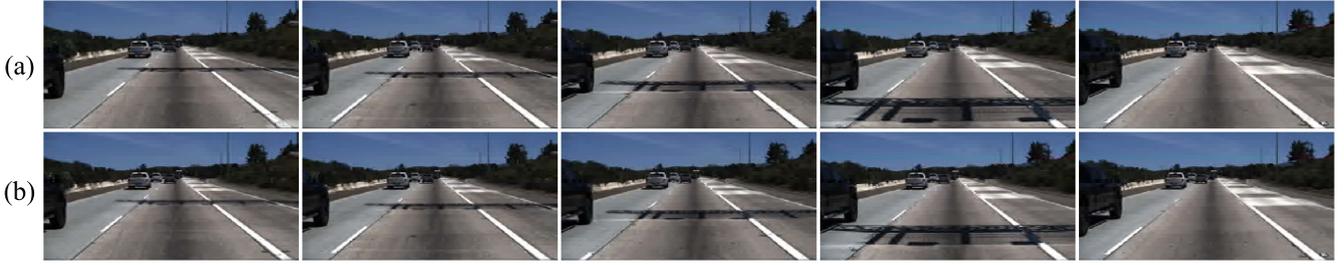

Fig. 5. Sample input frames of different time intervals, the details of which are shown in Table II. (a) same interval (0.15$s$), and (b) different intervals.

## IV. EXPERIMENTS

In this section, the robustness and accuracy of our proposed lane detection model are verified using extensive experiments. We qualitatively and quantitatively compare several state-of-the-art deep learning models, such as PINET(32 × 16) [26], PINET(64 × 32) [26], Res18-Qin [29], Res34-Qin [29], SCNN [31], LaneNet [28], U-Net [36], SegNet [46], U-Net_ConvLSTM [4] and SegNet_ConvLSTM [4], with our model.

### A. Dataset

*1) TuSimple Dataset:* The TuSimple lane marking challenge dataset [47] is used for training and testing our network. The TuSimple dataset consists of 3,626 training and 2,782 testing clips that are captured at different time periods under different weather conditions. There are 20 continuous frames in each one-second clip. Only the lane lines in the $20^{th}$ frame are officially marked as the ground truth in each clip. There are 2-5 lane lines in each labeled frame as the driving environment scenes change due to the different conditions on different sections of road, such as crossings, intersection merging and splitting, curved lane, and so on. The $K$ continuous frames ($K = 5$ in this work) of each clip are used as the input to train our model, and the output of $K^{th}$ frame is used to identify the lane lines by comparing it with the fixed label frame. The spatio-temporal information of previous K frames is extracted by double ConvGRUs to predict the lanes of $K^{th}$ frame. We sample $K$ different frames from each clip to construct new datasets in order to verify the performance of the proposed method when using different spatio-temporal information. Table II shows the different frames from each clip at different time intervals. Dataset $1^{\#}$ has the same time interval between adjacent frames. Dataset $2^{\#}$ has different time intervals between continuous frames. Some sample images of the two datasets are given in Fig. 5, we can clearly see that different frame sequences contain different information.

*2) Unsupervised LLAMAS Dataset:* Another dataset used for training and testing is Unsupervised LLAMAS dataset released by [48]. It is one of the largest public lane marker datasets and consists of 100,042 images. Among them, 79,113 images are used for training and 20,929 images are used for testing. Because of that there are no corresponding labels for 20,929 images, we divide 79,113 into two groups in our experiments, 58,269 images are used for training and 20,844 images are used for testing. Unsupervised LLAMAS dataset is different from TuSimple dataset, its characteristics are as follows: (1) the ground truth of lane lines are automatically labeled by the software. (2) the number of pixels occupied by the marked lane lines varies according to its distance and location. (3) the marked lane lines are intermittent and sporadically distributed. (4) for each image, the number of positive pixels is very small, about 2%. These characteristics may pose a great challenge to the models that are trained and tested on it.

TABLE II
RECONSITUTION DATASET FROM ORIGINAL TUSIMPLE DATASET

| Dataset | Labeled | Sample | Time space($s$) |
|---|---|---|---|
| 1 | $20^{th}$ frame | $1^{th}, 5^{th}, 10^{th}, 15^{th}, 20^{th}$ | 0.15 |
| 2 | $20^{th}$ frame | $2^{th}, 5^{th}, 9^{th}, 14^{th}, 20^{th}$ | 0.15,0.2,0.25,0.3 |

TABLE III
RECONSITUTION DATASET FROM ORIGINAL UNSUPERVISED LLAMAS DATASET

| Dataset | Labeled | Sample | Train | Test |
|---|---|---|---|---|
| 1 | last frame | $1^{th}, 2^{th}, 3^{th}, 4^{th}, 5^{th}$ | 11,650 | 4,168 |
| 2 | last frame | $1^{th}, 3^{th}, 6^{th}, 10^{th}, 15^{th}$ | 3,880 | 1,389 |

We build two new datasets according to the original Unsupervised LLAMAS dataset. Table III shows more information. Dataset $1^{\#}$ has the same time interval between frames and each consecutive five pictures is taken as a record in it. In dataset $1^{\#}$, 11,650 records are used for training and 4,168 records are used for testing. Dataset $2^{\#}$ has different time intervals between frames and each consecutive fifteen pictures is taken as a record in it. In dataset $2^{\#}$, 3,880 records are used for training and 1,369 records are used for testing.

### B. Qualitative Evaluation

*1) TuSimple Dataset:* A qualitative evaluation is the most intuitive analysis for evaluating the performance of one algorithm. In Fig. 6, we present some lane detection results of our model and other state-of-the-art deep learning models in the challenging scenes. All of the results from the different models are not post-processed. Those challenging scenes displayed in Fig. 6 are quite often the combinations of multiple scenarios, as shown in Table IV. We detail the challenging scenes including broken lines, solid lines, unclear broken lines, yellow lines, no lines, occlusions, curves, intersections and splits in the left and right-side roads, respectively, in Table IV. For example, the first input frame shown in Fig. 6 includes a solid line, a shadow, and a broken line. We hold that all the properties of the lane lines and driving environment scenes should be considered when comparing the performance of each model. The robust algorithms should work well in different challenging scenes.

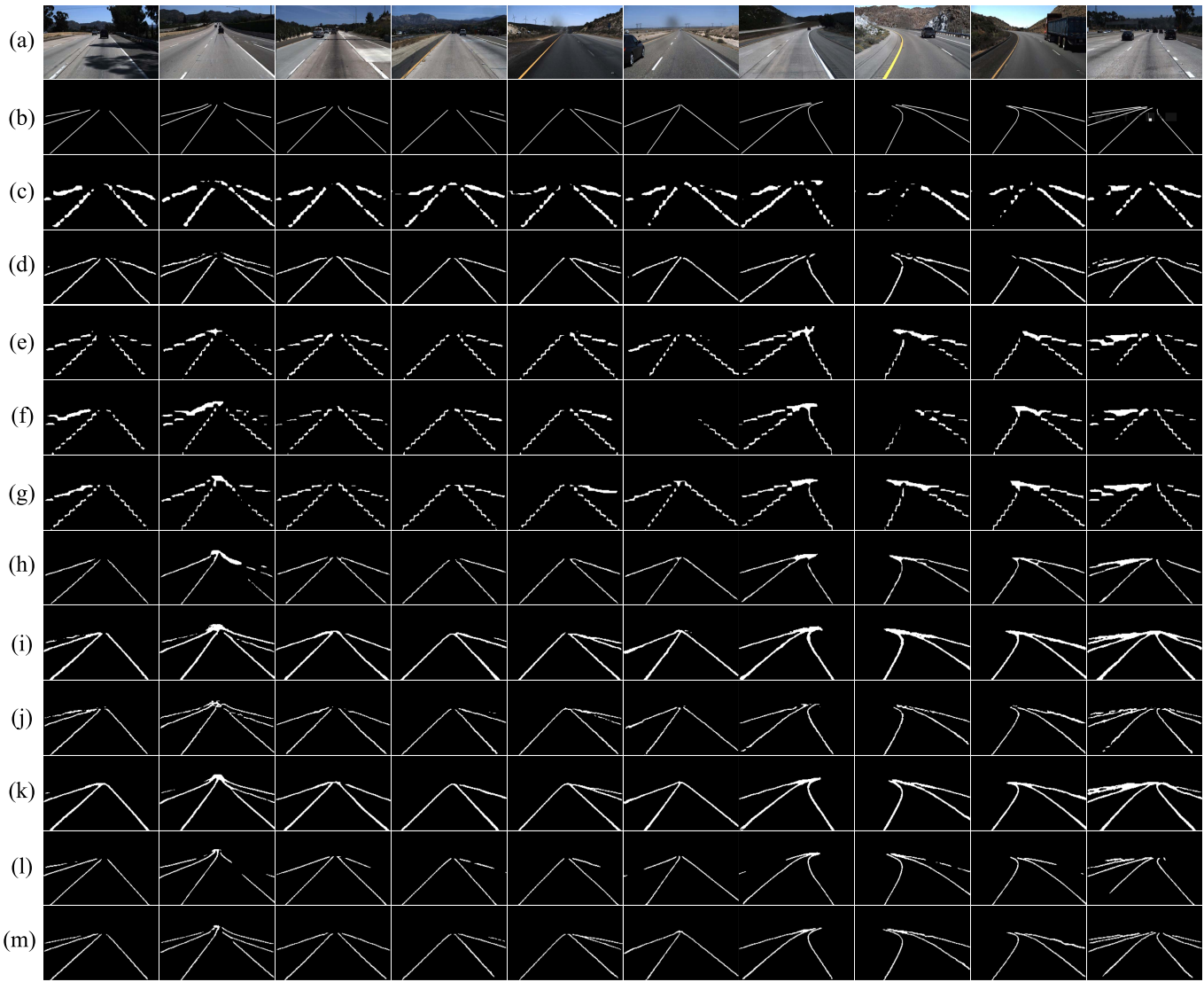

Fig. 6. Qualitative evaluation of our proposed model and other state-of-the-art deep learning models on TuSimple dataset. All of the detection results are not postprocessed. (a) Input frames. (b) Ground truth. (c) PINET(32 × 16). (d) PINET(64 × 32). (e) Res18-Qin. (f) Res34-Qin. (g) SCNN. (h) LaneNet. (i) SegNet. (j) SegNet_ConvLSTM. (k) U-Net. (l) U-Net_ConvLSTM. (m) Proposed model.

TABLE IV
ROAD CONDITIONS IN EACH INPUT FRAME SHOWN IN FIG. 6

| Input frames # | Road conditions in each input frame | |
| --- | --- | --- |
| | Left-side of the road | Right-side of the road |
| 1 | Broken line | Shadow, solid line |
| 2 | Unclear broken line | Solid line, intersection, broken line |
| 3 | Broken line | Interference signal, solid line |
| 4 | Yellow line, no line | Solid line, splitting |
| 5 | Yellow line | Unclear broken line, no line |
| 6 | Broken line, occlusion | Solid line |
| 7 | No line, broken line | Solid line, curve |
| 8 | Yellow line, curve | No line, occlusion, curve |
| 9 | Yellow line, curve | Broken line, occlusion |
| 10 | No line, broken line | Broken line, solid line, intersection |

The properties of lane lines may mainly include the following aspects:

- The number of lane lines should not be mispredicted or misjudged. A wrong detection or missed detection may lead to the consequence that the self-driving car considers unreachable areas as drivable areas. If that occurs, there is no doubt that such an incorrect prediction would surely results in unimaginable consequences. As shown in the first and second columns of Fig. 6, the proposed model can clearly detect the number of lane lines. However, other models more or less cannot detect the complete lane lines.

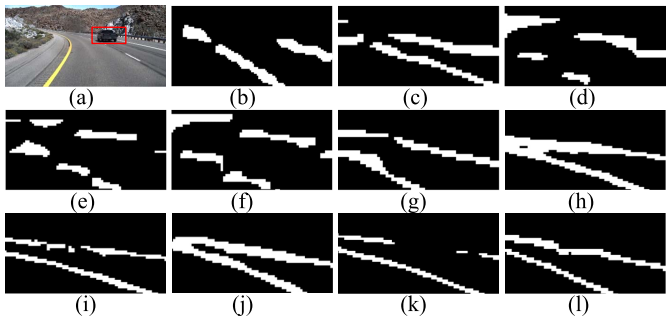

Fig. 7. The details of lane detection results when occlusion exists. (a) Input frames. (b) PINET(32 × 16). (c) PINET(64 × 32). (d) Res18-Qin. (e) Res34-Qin. (f) SCNN. (g) LaneNet. (h) SegNet. (i) SegNet_ConvLSTM. (j) U-Net. (k) U-Net_ConvLSTM. (l) Proposed model.

- The location of each lane line should be exactly the same as the corresponding ground truth. It can be seen from Fig. 6 that, compared with other models, the results of the proposed model (Fig. 6(m)) are almost identical to those of the corresponding ground truth (Fig. 6(b)). In Fig. 6, U-Net_ConvLSTM either does not predict the locations of lane lines or partly obtains a few lines relative to the ground truth in the challenging scenes. The same problems occur in PINET(32 × 16), PINET(64 × 32), Res18-Qin, Res34-Qin, SCNN, LaneNet, SegNet_ConvLSTM, U-Net, and SegNet in their own corresponding predicted images.
- The continuity of lane lines requires an unbroken prediction from the start to the end of the lane line. If there are some discontinuous lane lines in the prediction, a self-driving vehicle may mistake the missing part as a drivable area. In Fig. 6, we can see that most of the lane prediction results from other models are discontinuous. Compared to the results from our model in Fig. 6, our results have good continuity, which is the same as the ground truth.
- The driving environment scene should be considered when qualitatively evaluating the performance of one algorithm. Apparently, an algorithm not only performs well in common scenes but, more importantly, also performs well in challenging scenarios that often have higher requirements for models because those scenes are often combinations of different driving scenes. In Fig. 6, the conditions in each input frame are different from the others, which are listed in Table IV. Experiments show that the proposed algorithm can accurately and completely detect all the lines even in the challenging scenarios. Furthermore, other models more or less lose something in the challenging scenes. Taking the occlusion in the eighth column of Fig. 6 as an example, our proposed method accurately detects the lines. However, either a part of the lines or rougher lines are detected by the other models. The details of the prediction results are shown in Fig. 7. We can see that the prediction lane results of other models are rougher or missing.

*2) Unsupervised LLAMAS Dataset:* As shown in Fig. 8, for those models whose output results are discrete points, including PINET(32 × 16), PINET(64 × 32), Res18-Qin and Res34-Qin, their detection results are still not good. For the

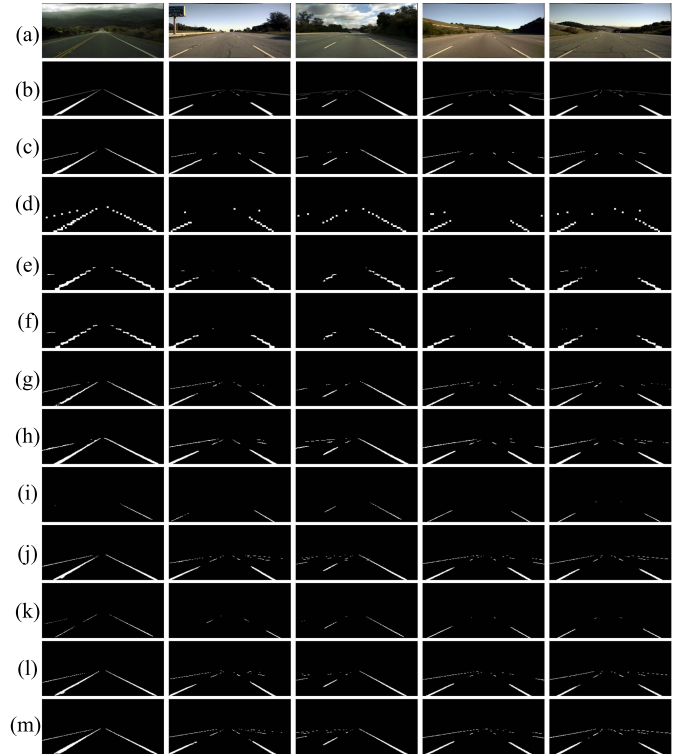

Fig. 8. Qualitative evaluation of our proposed model and other state-of-the-art deep learning models on Unsupervised LLAMAS dataset. All of the detection results are without postprocessing. (a) Input frames. (b) Ground truth. (c) PINET(32 × 16). (d) PINET(64 × 32). (e) Res18-Qin. (f) Res34-Qin. (g) SCNN. (h) LaneNet. (i) SegNet. (j) SegNet_ConvLSTM. (k) U-Net. (l) U-Net_ConvLSTM. (m) Proposed model.

Unsupervised LLAMAS dataset with fewer labeled pixels, the performance of U-Net and SegNet is no longer the same as they are on TuSimple dataset, and they detect only a small number of lane lines. SCNN, LaneNet, SegNet_ConvLSTM, U-Net_ConvLSTM and our proposed model can detect most lane lines. But for the detection results, as far as the width, location and number of lane lines, our proposed model is better than other models.

### C. Quantitative Evaluation

*1) TuSimple Dataset:* A quantitative evaluation is usually used for evaluating the performance of algorithms in computer vision research. To accurately and comprehensively evaluate the algorithms, we choose four evaluation metrics to evaluate our proposed method: Accuracy, Precision, Recall and F1-Measure. Accuracy shown in Formula 5 involves the correct judgment of pixels and is quite often used for evaluating the overall performance [49], [50]. True Positive means that the forecasted results are consistent with actual results. Here, True Positive represents the number of lane pixels that are correctly predicted as lanes. Similar meanings exist for True Negative, False Positive and False Negative. True Negative denotes that a pixel's value in the predicted image equals the corresponding background value, which means that the corresponding pixel point does not belong to any lane lines.

$$Accuracy = \frac{True\ Positive + True\ Negative}{Total\ Number\ of\ Pixels} \quad (5)$$

TABLE V
PERFORMANCE COMPARISON OF OUR MODEL WITH STATE-OF-THE-ART MODELS ON THE TUSIMPLE DATASET

| method | Accuracy(%) | Precision | Recall | F1-Measure | Params(M) |
| --- | --- | --- | --- | --- | --- |
| PINET(32x16) | 95.52 | 0.504 | 0.656 | 0.570 | 17.9 |
| PINET(64x32) | 97.62 | 0.847 | 0.895 | 0.871 | 17.9 |
| Res18-Qin | 96.90 | 0.630 | 0.691 | 0.659 | 58.4 |
| Res34-Qin | 96.89 | 0.637 | 0.701 | 0.668 | 98.9 |
| SCNN | 96.79 | 0.654 | 0.808 | 0.722 | 76.9 |
| LaneNet | 97.94 | 0.875 | 0.927 | 0.901 | 78.8 |
| SegNet | 96.05 | 0.730 | 0.981 | 0.838 | 117.9 |
| SegNet_ConvLSTM | 97.96* | 0.852 | 0.964 | 0.901 | 268.9 |
| U-Net⋆ | 97.96 | 0.864 | 0.955 | 0.908 | 4.99 |
| U-Net_ConvLSTM | 98.22* | 0.857 | 0.958 | 0.904 | 204.6 |
| Proposed† | 98.04 | 0.8750 | 0.9531 | 0.9124 | 13.4 |
| Proposed‡ | 97.98 | 0.8674 | 0.9562 | 0.9097 | 13.4 |

*: average value calculated by two values from Test_acc in [4].
U-Net⋆: the revised U-Net shown in Table I.
Proposed† and Proposed‡: Trained on dataset $1^{\#}$ and $2^{\#}$ in Table II, respectively.

As seen from Table V, the accuracy value of proposed model is very close to that of U-Net_ConvLSTM, which is the most accurate one, but the proposed model is still more accurate than the others. Accuracy is a reference metric for evaluating the lane detection task, and its value initially reflects the rationality of the proposed network model. In our proposed network model, we treat lane detection as a binary classification task. The number of pixels belonging to the background is far greater than the number belonging to the ground truth in the dataset. Therefore, we may obtain a higher accuracy than the one in Table V by adjusting the weights of both the background and ground truth when training and testing the model. However, for an algorithm, the wrong judgment of pixels cannot be ignored. Precision and Recall are two other metrics that reflect a fairer and more reasonable comparison when calculating the performance of an algorithm [4], [31]. They are defined as

$$Precision = \frac{True\ Positive}{True\ Positive + False\ Positive} \quad (6)$$

$$Recall = \frac{True\ Positive}{True\ Positive + False\ Negative} \quad (7)$$

False Positive represents the number of pixels belonging to the background that are wrongly predicted as the ground truth. False Negative represents the number of pixels belonging to the ground truth that are predicted as background. As exhibited in Table V, the Precision of our proposed model is very close to the highest value of LaneNet. The Recall of our proposed model is slightly lower than that of SegNet_ConvLSTM. Combined with the qualitative evaluation, the higher Precision is mainly due to the following predicted items: (a) the number of lane lines, (b) the position of each lane line, and (c) the continuity of the lane lines. The correct prediction directly reduces those misjudged operations which increases the number of False Positives. With respect to the model's structure, a further explanation of the high Precision and Recall of our proposed model can be summarized as follows: First, the nonlinear ConvGRU makes our model have a better information extracting ability from its input vector. Second, the FCGRU can effectively extract and remember the low-level features shown in Fig. 4. The further extracted features can compensate for the missing information caused by the convolutional operations. Finally, the MCGRUs make full use of the spatio-temporal information by processing $K$ frames and adjusting the relative weights of the convolutional kernels.

The F1-Measure, as a benchmark that balances Precision and Recall, is also used to evaluate the performance of our proposed model. The F1-Measure metric is defined as

$$F1 - Measure = \frac{2 * Precision * Recall}{Precision + Recall} \quad (8)$$

In the F1-Measure, the weights of Precision and Recall are equal. The experimental results show that the F1-Measure is directly related to the performance of the algorithm. As shown in Table V, the F1-Measure of our proposed model is the highest among all the models. The experiments show that it is impossible to increase the F1-Measure by increasing only the Precision or Recall. Only when both Precision and Recall increase will the F1-Measure increase. Therefore, although U-Net, SegNet and SegNet_ConvLSTM have the higher Recalls or LaneNet has the higher Precision, their F1-Measures are lower than our mode's F1-Measures. From the above analysis, regardless of whether we conduct a qualitative evaluation or a quantitative evaluation, our proposed model outperforms the other state-of-the-art deep learning models. We can conclude that the proposed model can detect the lanes more precisely than other models under the conditions of challenging scenes. Furthermore, we further compare the performance of our proposed model trained on dataset $1^{\#}$ and dataset $2^{\#}$. As shown in the last two rows of Table V, the scores on dataset $1^{\#}$ are slightly higher than those on dataset $2^{\#}$. Therefore, we consider that the same time interval between the input continuous frames leads to more favorable results than using different time interval for the lane detection task in our proposed model.

*2) Unsupervised LLAMAS Dataset:* Besides Precision and Recall, the Formula 9 is also used to evaluate the models which are trained and tested on Unsupervised LLAMAS dataset. The Formulas 6 and 7 provide more information about Precision and Recall.

$$AP = \sum_{p=1}^{U}(\sum_{q=1}^{V+1}(Precision_q * \Delta Recall_q)) \quad (9)$$

The variable $AP$ means average precision, $U$ means the number of all test images. $V$ means the number of samples for a single image. $\Delta Recall$ represents a difference between adjacent samples for the values of $Recall$. The variables $p$ and $q$ are subscripts. In actual calculation, $Recall_0$ is set to 0, $Precision_0$ is set to 1, and the variable $V$ is set to 100.

As can be seen from Table VI, our proposed model achieves very competitive results. The AP values of our proposed model are close to the highest value of U-Net, but the

TABLE VI
PERFORMANCE COMPARISON OF OUR MODEL WITH STATE-OF-THE-ART MODELS ON THE UNSUPERVISED LLAMAS DATASET

| Method | AP | Precision | Recall |
|---|---|---|---|
| Simple Baseline [49] | 0.4340 | 0.5460 | 0.4500 |
| GAC Baseline1* | 0.7780 | 0.7480 | 0.3070 |
| PINET(32x16) | 0.8193 | 0.5825 | 0.5940 |
| PINET(64x32) | 0.8348 | 0.6200 | 0.6020 |
| Res18-Qin | 0.6532 | 0.4635 | 0.4056 |
| Res34-Qin | 0.6546 | 0.4574 | 0.4052 |
| SCNN | 0.8231 | 0.5754 | 0.6033 |
| LaneNet | 0.8179 | 0.4193 | 0.7022 |
| SegNet | 0.8959 | 0.8735 | 0.2102 |
| SegNet_ConvLSTM | 0.8500 | 0.5487 | 0.6839 |
| U-Net | 0.9145 | 0.8861 | 0.2965 |
| U-Net_ConvLSTM | 0.8510 | 0.5857 | 0.6558 |
| Proposed† | 0.8540 | 0.6005 | 0.6501 |
| Proposed‡ | 0.8519 | 0.6162 | 0.6313 |

GAC Baseline1*: Not public yet according to the official announcement. Proposed† and Proposed‡: Trained on dataset 1# and 2# in Table III, respectively.

corresponding value of Recall of U-Net is very low, less than 0.3. As mentioned earlier, the lower the value of Recall, the higher the value of FN, which means the more positive points are considered as the negative points. Compared with other models, the evaluation metrics calculated by our proposed model always keep a balance. When the values of Precision and Recall are synchronously improved in our proposed model, the difference between them is always kept in a certain range, while the corresponding difference between Precision and Recall from other models fluctuates greatly. For example, the values of AP from SegNet_ConvLSTM and U-Net_ConvLSTM are closer to our proposed model, but the difference between Precision and Recall from them is 0.14 and 0.07, respectively, while the difference from our proposed model is 0.05 (0.015).

These experimental results further verify the calculation Formulas, and further explain that the output of a model is closer to the ground truth only when all the relevant evaluation metrics are improved and the difference between them is small and within a certain range.

### D. Ablation Study

We investigate the effects of the model with only MCGRUs (shown in Table VII) and also perform extensive experiments to investigate the effects of different locations of FCGRU (shown in Table VIII), e.g., embedding FCGRU into the low level layers (such as conv1_2) or the high-level layers (such as conv5_2).

The performance of the model with only MCGRUs is discussed in Table VII. The experimental results show the following: (1) the performance of same time interval is better than the different time interval. The reason may be that equal interval between frames makes the missing information present regularity and stability, which enables ConvGRUs to look back in the past better and predict the future frame well. However, the sequential and unequal interval between frames fluctuates greatly and destroys the regularity and stability of information. (2) under the same conditions, the performance of the model with FCGRU and MCGRUs is better than that of the model with only MCGRUs. The possible reason is that FCGRU can try its best to memorize and retain the most likely features of lane lines (as shown in Fig. 4 (b)).

We have discussed the experimental results of different locations of FCGRU in our proposed model, as shown in Table VIII. The experimental results show the following: (1) When trying to embed FCGRU into the lowest level layer (such as conv1_2), the performance of the corresponding model is not ideal. The possible reason is that the low-level feature layer mainly contains local information. (2) When trying to embed FCGRU into the highest level layer (such as conv5_2), the performance of the corresponding model is not ideal. The possible reason is that the highest-level feature layer mainly contains global information. (3) When trying to embed FCGRU close to a middle-level layer (such as conv2_2), the performance of the corresponding model is ideal. The possible reason is that the FCGRU may act as a connector between the local and global information. Therefore, we set the FCGRU at a layer (after conv2_2) in our work.

## V. DISCUSSION

### A. The Drawbacks of Ground Truth

There are some problems encountered in the experiment that may directly affect the evaluation results of our model. One problem is due to the official lane labels. In some cases, the official lane labels do not exactly match the actual conditions in the TuSimple dataset. For example, as shown by the second, third and last rows in Fig. 9, the number of lane lines marked in the testing dataset does not match the number of lane lines that obviously exist in the original images. Such labels, which are inconsistent with the actual conditions, undoubtedly cause some of the ground truth to be treated as background during the testing process. Nevertheless, the proposed model detects those extra lanes.

The paradox stated above directly leads to more False Negative and a lower Recall. It may aggravate the changes of the internal weights which further make the proposed network difficult to stabilize when training and testing the model. Nevertheless, our proposed network can still detect the lane lines shown in the second, third, and fourth rows of Fig. 9. However, according to the facts, the extra predicted results are considered to be False Negative pixels when calculating the metrics. Obviously, for an algorithm, this reduces the values of the evaluation metrics.

Another problem is that the lengths of the labeled ground truth are shorter than the actual lengths in the original images. However, the proposed model can detect more information than the marked length as shown in the first row of Fig. 9. Nevertheless, these extra predicted parts are also considered to

TABLE VII
PERFORMANCE COMPARISON OF DIFFERENT MODULE IN PROPOSED NETWORK ON TUSIMPLE TEST DATASET

| Method | Dataset 1 | | | | Dataset 2 | | | |
|---|---|---|---|---|---|---|---|---|
| | Acc(%) | Pre | Rec | F1-M | Acc(%) | Pre | Rec | F1-M |
| FCGRU | 98.05 | 0.8791 | 0.9419 | 0.9094 | 97.97 | 0.8643 | 0.9563 | 0.9082 |
| MCGRUs | 98.00 | 0.8668 | 0.9561 | 0.9093 | 97.96 | 0.8656 | 0.9564 | 0.9087 |
| FCGRU and MCGRUs | **98.04** | 0.8750 | 0.9531 | **0.9124** | **97.98** | 0.8674 | 0.9562 | **0.9097** |

Dataset 1 and Dataset 2 are shown in Table II.
FCGRU: only with FCGRU module. MCGRUs: only with MCGRUs module. FCGRU and MCGRUs: with both FCGRU and MCGRUs module.

TABLE VIII
PERFORMANCE COMPARISON OF DIFFERENT LOCATIONS OF FCGRU IN PROPOSED NETWORK ON TUSIMPLE DATASET

| Location | Dataset 1 | | | | Dataset 2 | | | |
|---|---|---|---|---|---|---|---|---|
| | Acc(%) | Pre | Rec | F1-M | Acc(%) | Pre | Rec | F1-M |
| conv1_2 | 98.04 | 0.8864 | 0.9203 | 0.9035 | 97.87 | 0.8662 | 0.9421 | 0.9028 |
| conv2_2 | **98.04** | 0.8750 | 0.9531 | **0.9124** | **97.98** | 0.8674 | 0.9562 | **0.9097** |
| conv3_2 | 97.74 | 0.8373 | 0.9182 | 0.8772 | 96.65 | 0.7243 | 0.8565 | 0.7849 |
| conv4_2 | 97.96 | 0.8660 | 0.9474 | 0.9045 | 97.90 | 0.8611 | 0.9382 | 0.9010 |
| conv5_2 | 97.96 | 0.8622 | 0.9528 | 0.9053 | 97.93 | 0.8584 | 0.9545 | 0.9045 |

Dataset 1 and Dataset 2 are shown in Table II.
conv1_2: adding FCGRU after the convolutional operation of conv1_2 in Table I. Other meanings are similar.

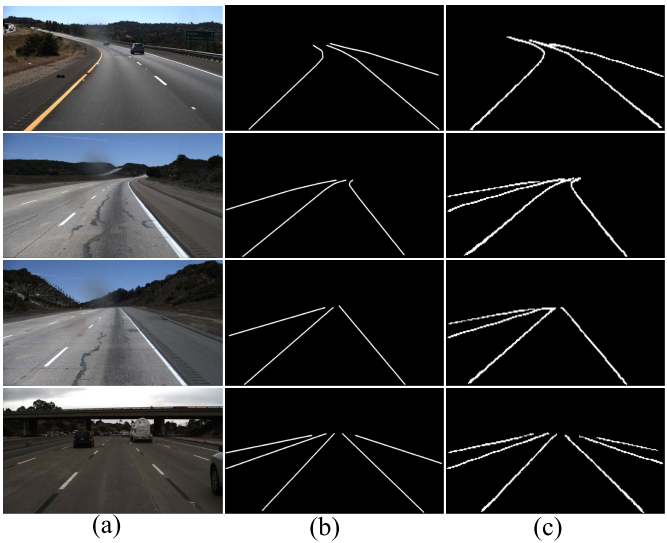

(a)      (b)      (c)

Fig. 9. Discussion of the labeled lanes in the TuSimple dataset. (a) Input frames. (b) Ground truth. (c) Prediction results of the proposed model.

TABLE IX
RESULTS ON TUSIMPLE DATASET WITH DIFFERENT $K$ CONTINUOUS FRAMES

| K values | Acc(%) | Pre | Rec | F1-M |
|---|---|---|---|---|
| K = 1 | 98.03 | 0.8760 | 0.9506 | 0.9118 |
| K = 2 | 98.02 | 0.8763 | 0.9512 | 0.9122 |
| K = 3 | 97.98 | 0.8688 | 0.9530 | 0.9089 |
| K = 4 | 98.05 | 0.8798 | 0.9470 | 0.9122 |
| K = 5 | **98.04** | 0.8750 | 0.9531 | **0.9124** |
| K = 6 | 98.03 | 0.8781 | 0.9493 | 0.9123 |
| K = 7 | 97.96 | 0.8651 | 0.9597 | 0.9099 |

be False Positive pixels when calculating the quantitative evaluation, which also decreases the values of the corresponding metrics to some extent.

*B. The Selection of K Continuous Frames*

In our work, $K$ continuous frames are input into the network, where $K$ is set to 5. Although the length of continuous frames is fixed to $K$, we have considered the different time intervals of input to make up for the lack of different $K$. Actually, $K = 5$ is a more appropriate value on the TuSimple and Unsupervised LLAMAS datasets caused by the selections of different time intervals. In addition, the different frames from each clip (7 values of $K$) at different time intervals also have been discussed to verify the performance of our model on TuSimple dataset. As shown in Table IX, we can see that the evaluation scores are the highest when $K = 5$, which indicates that $K = 5$ is more reasonable in our work.

## VI. CONCLUSION AND FUTURE WORKS

In this paper, a novel spatio-temporal network with double ConvGRUs is proposed for reliably detecting lane lines in the challenging scenarios. The spatio-temporal network includes encoder and decoder parts. The continuous frames are input into the encoder network. Considering that the low-level features extracted by the CNN contain much interference information and the advantages of the ConvGRU, a Front ConvGRU (FCGRU) is added to the encoder network to go a step further to extract features from the interfered low-level features. After that, we concatenate the filtered features from the FCGRU with the interfered features from the CNN together into the next layer of the encoder network. The outputs of the encoder network are input into the other Middle ConvGRUs (MCGRUs) to extract more effective features by dealing with the spatio-temporal information of multiple continuous frames. At last, the decoder network completes decoding using deconvolution and upsampling operations. Finally, the predicted results corresponding to $K^{th}$ frame are obtained. After training the model, we validate our network on the TuSimple and Unsupervised LLAMAS datasets, which are both designed for autonomous driving deep learning methods. Qualitative and quantitative evaluations are used to evaluate the results in the challenging conditions. The experimental results show that our proposed model outperforms other state-of-the-art deep learning models.

In this work, a lane detection model, which is an important research branch in autonomous driving, has been proposed. This work has the following potential positive impact in the society. This work may provide a performable lane detection technology and some driving decision assistance for autonomous driving cars or ADASs. At the same time, this work may have some negative consequences because the deep learning algorithm relies heavily on datasets, and our model is trained on two lane detection benchmarks that lack some weather conditions (such as fog, snow, rain and sandstorms). Furthermore, we should be cautious of the result of failure of the algorithm which could cause unsafe driving judgements or the occurrence of traffic accidents. In the future work, we will try to optimize the structure of proposed model to improve the lane detection accuracy. We will further consider improving the feature representation ability of our model, such as adding spatial pyramid pooling (SPP) to obtain multi-scale feature maps, or dilated convolutions to increase the receptive field to get more feature information of lane lines. Moreover, more lane detection datasets including complex driving scenes (such as nighttime, crowded road and bad weather) will be trained and tested by our model to improve the robustness of our proposed model. Some latest other complex driving scenes datasets, such as Vehicle Detection in Adverse Weather Nature (DAWN) [51], also could be developed to solve the lane detection problem.

ACKNOWLEDGMENT

The authors would like to thank Prof. Guoxiang Gu for his valuable comments and suggestions on this work.

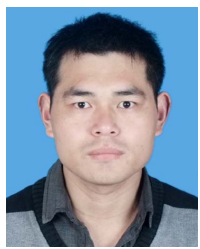

**Jiyong Zhang** is currently pursuing the Ph.D. degree with the School of Information Science and Technology, Southwest Jiaotong University, Chengdu, China.

His research interests include computer vision, deep learning, intelligent transportation, and embedded systems.

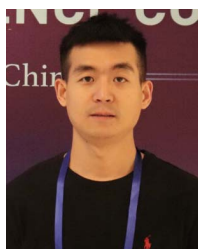

**Tao Deng** received the Ph.D. degree from the MOE Key Laboratory for NeuroInformation, University of Electronic Science and Technology of China, in 2018.

From 2016 to 2017, he was a Joint Ph.D. Student with the University of California, Santa Barbara, Santa Barbara, CA, USA. He is currently with the School of Information Science and Technology, Southwest Jiaotong University, Chengdu, China. His research interests include visual attention, cognition, computer vision, deep learning, image/video processing, and intelligent transportation.

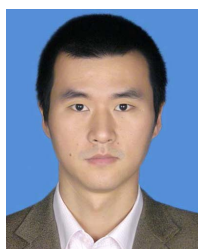

**Fei Yan** received the B.S. and M.S. degrees in mechanical engineering and automation from Northwestern Polytechnical University, China, in 2004 and 2007, respectively, and the Ph.D. degree in computer engineering from the Université de Technologies Belfort-Montbéliard, France, in 2012.

He is currently an Associate Professor with the School of Information Science and Technology, Southwest Jiaotong University, China. His research interests include control theory and engineering, artificial intelligence, and intelligent transportation systems.

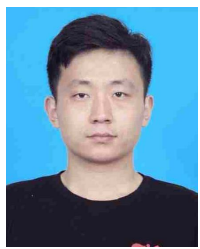

**Wenbo Liu** is currently pursuing the M.S. degree with the School of Information Science and Technology, Southwest Jiaotong University, Chengdu, China.

His research interests include computer vision, deep learning, and intelligent transportation.